\documentclass[sigconf]{acmart}

\usepackage{bbding}
\usepackage{balance}
\usepackage{fancyhdr}
\renewcommand\footnotetextcopyrightpermission[1]{}
\AtBeginDocument{%
  }

\acmISBN{978-1-4503-XXXX-X/18/06}

\acmSubmissionID{1631}

\begin{document}

\title{Generative Visual Commonsense Answering and Explaining with Generative Scene Graph Constructing}






\author{Fan Yuan\textsuperscript{\rm 1}, Xiaoyuan Fang\textsuperscript{\rm 1}, Rong Quan\textsuperscript{\rm 2}, Jing Li\textsuperscript{\rm 3}, Wei Bi, Xiaogang Xu\textsuperscript{\rm 4}, Piji Li\textsuperscript{\rm 1}$^{\ast}$}
\affiliation{%
  \institution{\textsuperscript{\rm 1} College of  Artificial Intelligence, Nanjing University of Aeronautics and Astronautics, \\ MIIT Key Laboratory of Pattern Analysis and Machine Intelligence, Nanjing, 211106, China \\
  \textsuperscript{\rm 2} College of Artificial Intelligence, Nanjing University of Aeronautics and Astronautics, \\
  the Key Laboratory of Brain-Machine Intelligence Technology, Ministry of Education, Nanjing, 211106, China \\
  \textsuperscript{\rm 3} Department of Computing, The Hong Kong Polytechnic University,  Hong Kong, China \\
  \textsuperscript{\rm 4} The Chinese University of Hong Kong, Hong Kong, China \\
  }
  \city{}
  \country{}}
\email{{fanyuan, pjli}@nuaa.edu.cn}

\renewcommand{\shortauthors}{Yuan et al.}

\begin{abstract}
  Visual Commonsense Reasoning, which is regarded as one challenging task to pursue advanced visual scene comprehension, has been used to diagnose the reasoning ability of AI systems. However, reliable reasoning requires a good grasp of the scene's details. Existing work fails to effectively exploit the real-world object relationship information present within the scene, and instead overly relies on knowledge from training memory. Based on these observations, we propose a novel scene-graph-enhanced visual commonsense reasoning generation method named \textit{\textbf{G2}}, which first utilizes the image patches and LLMs to construct a location-free scene graph, and then answer and explain based on the scene graph's information. We also propose automatic scene graph filtering and selection strategies to absorb valuable scene graph information during training. Extensive experiments are conducted on the tasks and datasets of scene graph constructing and visual commonsense answering and explaining, respectively. Experimental results and ablation analysis demonstrate the effectiveness of our proposed framework.
\end{abstract}




\maketitle

\renewcommand{\thefootnote}{\fnsymbol{footnote}}
\footnotetext[1]{Corresponding author.}
\renewcommand{\thefootnote}{\arabic{footnote}}

\section{Introduction}
Deep learning has greatly advanced in vision-language tasks, and has exhibited impressive performance, e.g., Image Captioning \cite{ref55, ref54}, ScienceQA \cite{ref53, ref21}. This advancement can help enhance the understanding and question-answering capabilities of web multimedia content, thereby more effectively processing and interpreting complex web visual information. Due to the black-box nature, models' explainability is receiving increasing attention. Visual Commonsense Reasoning (VCR) \cite{ref1} is a task that focuses on visual reasoning, requiring a model to simultaneously engage in reasoning from both recognition-level and cognition-level. At the same time, the Vision-language Natural Language Explanation (VL-NLE) \cite{ref11, ref12, ref15, ref14, ref13} task has extended traditional Natural Language Explanation (NLE) \cite{ref63, ref64, ref65} into the vision-language setting, aiming to help models generate clear answers and fine-grained explanations. Recently, VCR has also been incorporated into VL-NLE, transformed from a multiple-choice problem into a generation problem \cite{ref12, ref15, ref13}.

\begin{figure}[t]
    \centering
    \includegraphics[width=\columnwidth]{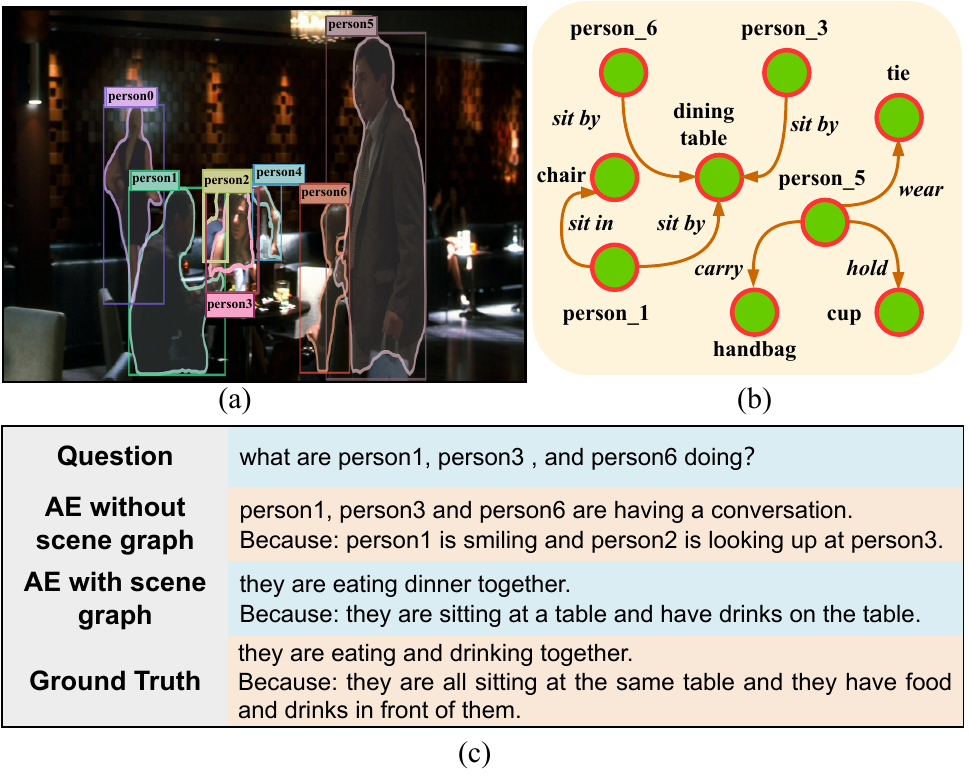}
    \vspace{-23pt}
    \caption{An example of visual commonsense reasoning. (a) is the image of the scene. (b) is a scene graph generated from (a). In the table of (c), the first row is the Question, the second row is the Answer and Explanation (AE) without scene graph, the third row is the Answer and Explanation (AE) with scene graph, and the fourth row is the Ground Truth.}
    \vspace{-20pt}
    \Description{..}
    \label{intro_image}
\end{figure}

Through observation, we find an encouraging phenomenon that a large number of specific objects and their relationships are contained in the questions, answers, and explanations. As shown in Table~\ref{object&relationship}, $99.7\%$, $97.6\%$, and $98.8\%$ of questions, answers, and explanations in the VCR dataset contain both the objects and their relationships. Each sentence of question, answer, and explanation in the VCR dataset contains more than 1 object and more than 2 relationships. In particular, each explanation contains an average of 4 objects and 3.8 relationships. For example, ``\emph{He is sitting in a cushioned piece of furniture. He has his briefcase on his lap and he is engrossed in a book.}'' contains relationships: ``\emph{He is sitting in furniture}'', ``\emph{briefcase on lap}'', and ``\emph{he is engrossed in a book}''. It is obvious that understanding the objects and their relationships well plays a crucial role in visual commonsense answering and explaining. Taking Figure~\ref{intro_image} as an example, when asked ``\emph{what are person1 and person3, and person6 doing?}'', a model that does not take object relationships into consideration generates the following: ``\emph{person1 and person3, and person6 are having a conversation. Because: person1 is smiling and person2 is looking up at person3.}''. However, in the ground truth explanation, critical factors such as ``\emph{dining table}'' and ``\emph{cup}'' explicitly contribute to identifying the scenario as a dining event rather than a mere conversational setting. It is evident that this prediction overly emphasizes the role of human factors and neglects other objects in the surrounding environment. Consequently, the statement generated with considering the object relationships ``\emph{they are eating dinner together. Because: they are sitting at a table and have drinks on the table.}'' is more precise.

\begin{table}[t!]
  \centering
  \scalebox{1.1}{
  \begin{tabular}{l|c|c|c}
    \toprule
          & Question & Answer & Explanation  \\
    \midrule
    \#Objects   & 1.2 & 2.0 & 4.0 \\

    \#Relations   & 2.1 & 2.1 & 3.8  \\
    \midrule
    Proportion  & $99.7\%$ & $97.6\%$ & $98.8\%$  \\
    \bottomrule
  \end{tabular}
     }
\caption{\textbf{A statistical analysis of the VCR dataset.} The second and third rows show the average number of objects and relations contained in each question, answer, and explanation. The fourth row shows the proportion of questions, answers, and explanations that contain goals and relationships, respectively.}
\vspace{-30pt}
\label{object&relationship}
\end{table}

Based on the above observations, we argue that current VCR models should additionally incorporate object relationships when making visual commonsense reasoning. Scene graph, as defined in the Scene Graph Generation (SGG) task \cite{ref5, ref18, ref17, ref16}, is known as a valuable means to model object relationships. A scene graph is a graph structure composed of nodes and edges, where each node represents an object, and each directed edge denotes the relationship between the objects at both ends. For example, as shwon in Figure~\ref{intro_image}~(b), in the triplet ``\emph{$\langle$person, holding, cup$\rangle$}'', ``\emph{person}'' and ``\emph{cup}'' are objects, and ``\emph{holding}'' is the relationship. By extracting various positional and semantic relationships among the objects, the potential of the scene graph has already been demonstrated in certain tasks \cite{ref9, ref10}. Considering that the images provided by VCR are insufficient to offer such detailed location and relationship information, the scene graph is well-suited to serve as a bridge to fill this gap during the reasoning process. 

A good scene graph is the key to the visual commonsense reasoning task. When constructing scene graphs, the most common approach is to first use an off-the-shelf detector such as Faster-RCNN \cite{ref35}) for object detection, followed by predicting relationships between the detected objects. However, the majority of contemporary SGG methods depend on object locations and bounding box data, which is not necessary for most downstream tasks \cite{ref57, ref58, ref59}, resulting in intricate mechanisms and pipelines. Taking the aforementioned shortcomings into account, we suppose that Large Language Models (LLMs), when endowed with visual understanding capabilities, can better distinguish the potential spatial relationships between two objects under a location-free setting.

To address the above-mentioned issues, we propose a framework named \textit{\textbf{G2}}, which represents for \textit{\textbf{G}}enerative visual commonsense answering and explaining with \textit{\textbf{G}}enerative scene graph constructing. In order to obtain more comprehensive visual information, we use the patch sequence from CLIP \cite{ref56} as input. To achieve a  more reliable perception of reasoning, we first train our model with LLMs on the Visual Genome (VG) dataset \cite{ref5} and generate a series of scene graphs for the VCR dataset. Meanwhile, to identify the most advantageous scene graphs for VCR generation, we also develop an automated selection mechanism, which assigns weights to the input based on the confidence score of the scene graph, thereby affecting the subsequent attention computation. This mechanism enables the model to seamlessly sift through highly reliable and valuable scene graphs throughout the generation process. During the VCR generation process, we implement an early fusion between the image patches and the text input. 
Extensive experimental results on the VCR dataset demonstrate our method outperforms the strong baselines with both automatic and human evaluation metrics. To further validate the performance, we conduct experiments on VQA-X \cite{park2016attentive} and e-SNLI-VE \cite{ref12}. The results provide additional confirmation of the effectiveness of our method.

To summarize, our contributions are three-fold:
\begin{itemize}
    \item We systematically explore how to combine patch sequences and LLMs for location-free scene graph construction, then utilize the explicit information of scene graph triplets for generating answers and explanations.

    \item We propose an automatic scene graph selection method at the VCR generation stage, which introduces confidence scores of scene graphs to weigh the input. The empirical results suggest the automatic selection approach has the best performance, compared to threshold-based approaches.

    \item Extensive experimental results show that our proposed \textit{\textbf{G2}} framework is superior to the strong baselines and is capable of generating more well-founded and rational explanations.
\end{itemize}


\begin{figure*}[t]
  \centering
  \includegraphics[width=1.0\textwidth]{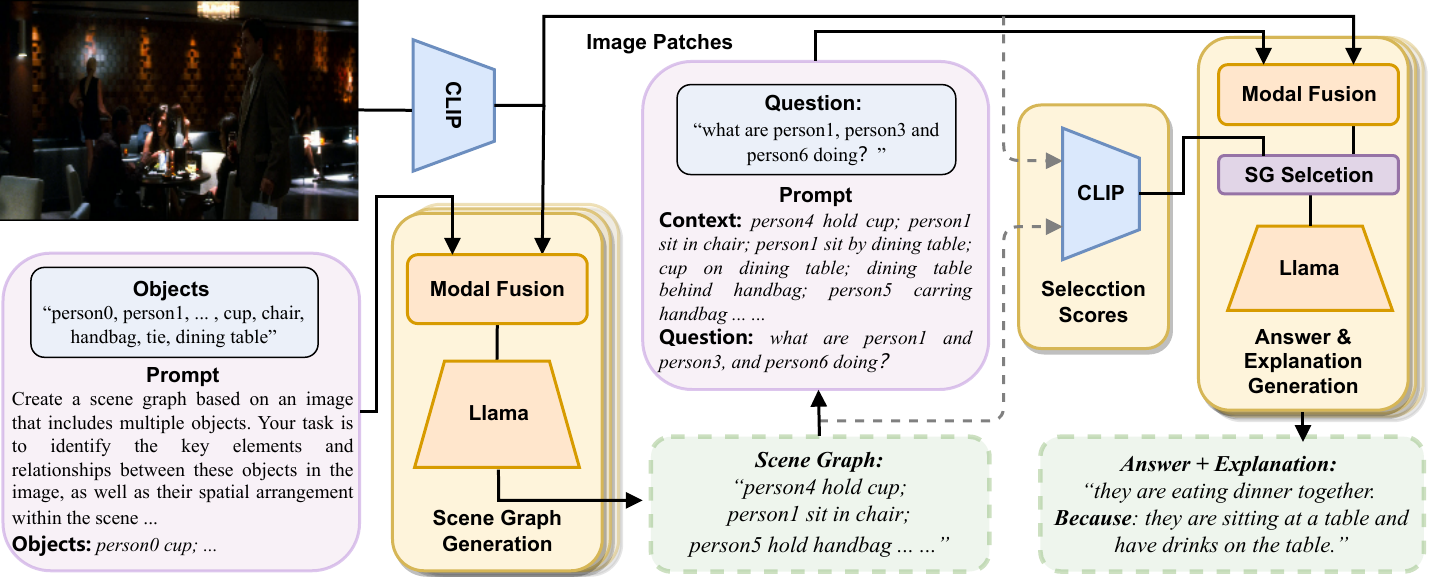}
  \vspace{-15pt}
  \caption{\textbf{An overview of our proposed \textit{\textbf{G2}}}. It first generates a scene graph based on the patch sequence and object prompt of the image. Then, combined with the question and image, it automatically selects the scene graph during training, and then generates answers and explanations that are consistent with commonsense.}
  \Description{..}
  \label{methods}
\vspace{-5pt} 
\end{figure*}

\section{Related Work}
\label{sec:related_work}

\subsection{Scene Graph Generation}
A scene graph contains semantic information detailing the objects, attributes, and relationships. To be specific, it is a graph structure composed of nodes and edges, where each node represents an object, and each directed edge denotes the relationship between the nodes at both ends. The predominant representation of scene graphs is triplets, in which the subject and object embody the endpoints, while the relationship signifies the directed edge. The Scene Graph Generation task (SGG) is to extract such semantic representations from images. Current approaches can be divided into two categories. The first category comprises two stages: first, object detection, such as using Faster-RCNN \cite{ref35}, followed by relationship prediction \cite{ref32, ref33, ref34}. Considering that elements in an image serve as the context for others, the second method jointly predicts objects and object relationships \cite{ref36, ref37, ref38}. Some work has also focused on the long-tail problem in scene graphs \cite{ref39, ref40, ref41} and proposed unbiased SGG \cite{ref42, ref43}. In addition, aiming to generate without using location information such as bounding boxes, location-free SGG \cite{ref2} has recently been proposed. Although existing work has been able to generate relatively high-quality scene graphs, there is still a long way to go before it becomes truly practical. In light of this situation, we want to utilize LLMs' powerful understanding ability for generation without positional information.
\subsection{Visual Commonsense Reasoning}
Visual Commonsense Reasoning \cite{ref1} is a recently proposed multi-modal task. The VCR dataset provides a large number of images and related questions, requiring answers and explanations to be inferred based on them. Each piece of data contains four alternative answers and four alternative reasoning options. Unlike the classic Visual Question Answering task, the options provided by VCR usually involve a comprehensive description of a set of object relationships or an event. Since it necessitates a thorough grasp of scene information and object relationships, as well as a profound understanding of the commonsense information provided by the images, making the correct choice simultaneously is challenging. In the past few years, many works have adopted different methods. R2C \cite{ref1} initially proposes a three-stage pipeline model. Later, much work adopts different pre-training models \cite{ref44, ref45, ref46, ref49} to solve the deep understanding problems of VCR. At the same time, some work attempts to use knowledge distillation \cite{ref47} and adversarial training \cite{ref48}, aiming to enhance performance with additional knowledge. However, these works only limit this reasoning approach to classification problems, making it challenging to apply to more general application scenarios.

\subsection{Vision-language Natural Language Explanation}
VL-NLE is an extension of Natural Language Explanation in the multi-modal domain. It aims to explain the decision-making process of black-box models by generating natural language sentences that are human-friendly and fine-grained. The e-ViL \cite{ref12} is a recent evaluation benchmark for VL-NLE tasks, which contains three datasets, including VQA-X \cite{ref52}, VCR, and e-SNLI-VE. It should be noted that this benchmark transforms the VCR task from a multiple-choice problem into a generation problem. So far, much work \cite{ref11, ref12, ref13, ref14, ref15} has been devoted to improving the explanation ability of models. For example, The e-UG \cite{ref12} uses UNITER \cite{ref49} as a prediction module to make explanation prediction, NLX-GPT \cite{ref14} combines answer and explanation generation by training a distilled GPT-2. While existing methods still have difficulty capturing accurate object relationships in the scene to facilitate reasoning, and this is the problem our work attempts to address.


\section{The Proposed G2 Framework}
\subsection{Overview}
\label{formulation}
The task investigated in this work is to generate reasonable answers and explanations for Visual Commonsense Reasoning (VCR) based on the input images and questions, as well as the generated scene graphs from images. We illustrate our \textit{\textbf{G2}} framework in Figure~\ref{methods}. Let $I$ be the given image, $Q$ be the corresponding question, and $O = (o_1, o_2,\dots, o_n)$ be a set of objects, where $n$ indicates the number of objects. During the location-free scene graph constructing stage, take the patch sequence $P=(p_1, p_2,\dots, p_m)$ of image $I$ as the visual input $X_{\text{v}}$, and the prompt template containing objects as the text input $X_{\text{t}}$. The goal is to predict a scene graph $G = (V, E)$, where $V$ and $E$ represent entity nodes and relationships, respectively. Subsequently, when generating VCR answers and explanations, with an image $I$, a question $Q$, and a scene graph $G$.
The target is to generate the answer and explanation $(I,G,Q) \rightarrow (A, R)$, where $A$ represents the answer and $R$ denotes the explanation. We expect that the generated $A, R$, with the assistance of $G$ is reasonable and conforms to commonsense.

\subsection{Location-free SGG}
\label{sgg}
The location-free scene graph generation model is trained on the Visual Genome (VG) dataset. In VG, each image contains a different number of scene graphs, and the objects in these scene graphs also vary in size. In general perception, small objects, such as ``\emph{hair}'' and ``\emph{hand}'', can provide less information in their associated scene graphs, while larger objects often play a more critical role in inference. Therefore, we have performed a certain degree of cleaning on the VG dataset. We sort all scene graphs corresponding to each image according to the average size of objects and subjects and select the top 50 for training. These scene graphs can already include most of the valid objects and effectively exclude triples with less information. 

Since large language models cannot locate objects by location data such as bounding boxes, we do not use these data as input. In order to train the model to obtain the ability to generate scene graphs without bounding boxes, the object sets provided by the dataset are applied as part of the input with the necessary object information. However, in order to prevent the presence of similarly named objects in the image scene from affecting the prediction effect of the model, an ordinal number is given based on where the selected object appears from left to right in the image. For example, if there are multiple ``person'' objects in the scene, we define the first person on the left as ``person0'', the second as ``person1'', and so on. At the same time, we also add individual objects with no specific relationship to form a triplet of ``$\langle$subject, None, object$\rangle$'' to train the model to deal with irrelevant objects.

We use a decoder-only large language model Llama-3.2 \cite{llama3} as our backbone. Inspired by the recent success of Vision Transformers \cite{ref22, ref23}, we first employ an off-the-shelf CLIP \footnote{https://github.com/jianjieluo/OpenAI-CLIP-Feature} to extract a patch sequence of the image. Since our visual input is patch sequence, and Llama-3.2 is a text LLM, it is necessary to explicitly provide object information to help with modality alignment. Concretely, to get a predicted scene graph $G=\{sub_1\ obj_1\ rel_1; \ \dots\ ; sub_{\sigma}\ obj_{\sigma}\ rel_{\sigma}\}$, where $\sigma$ represents the number limit of generated scene graphs, we combine the subjects and objects from $Q$ into a textual input format $X_{\text{o}} = \{sub_1\ obj_1; sub_2\ obj_2;\ \dots\ ; sub_{\sigma}\ obj_{\sigma}\}$. As shown on the left side of Figure~\ref{methods}, we use the text prompt: ``[\emph{Create a scene graph based on an image that includes multiple objects. The task is to identify the key elements and relationships between these objects in the image, as well as their spatial arrangement within the scene. Objects:$\{X_{\text{o}}\}$ Scene:}]''. We take this prompt as text input $X_{\text{t}}$ and patch sequence $P$ as vision input $X_{\text{v}}$, then feed them into the scene graph generation model. 

In order to better help our model improve multimodal semantic understanding capability during training, we execute modality fusion prior to the Llama-3.2. So, contrary to previous work \cite{ref21}, we first apply a learnable projection matrix to project the visual feature into textual representation space before the LLM. Then a single-head cross-modal attention and gated mechanism \cite{ref24, ref25} are used to fuse the embeded vision input $\mathbf{H_{\text{v}}}$ and the embeded text input $\mathbf{H_{\text{t}}}$. We apply $\mathbf{H_{\text{t}}W^Q}$ as query, $\mathbf{H_{\text{v}}W^K}$ as key and $\mathbf{H_{\text{v}}W^V}$ as value for cross-attention calculation. Upon obtaining cross-modal embedding $\mathbf{H_{\text{c}}}$, we concatenate $\mathbf{W_t}\mathbf{H_{\text{t}}}$ with $\mathbf{W_v} \mathbf{H_{\text{c}}}$ and employ a sigmoid function to derive the gating parameter $\lambda$. Subsequently, the gating parameter $\lambda$ is utilized to fuse $\mathbf{H_{\text{t}}}$ and $\mathbf{H_{\text{c}}}$ for the fused embedding $\mathbf{H_{\text{f}}}$. This process can be represented by the following formulas:
\begin{equation}
     \mathbf{H_{\text{c}}} =  \boldsymbol{\mathrm{Attention}} \left(\mathbf{H_{\text{t}}W^Q}, \mathbf{H_{\text{v}}W^K}, \mathbf{H_{\text{v}}W^V}\right),
\end{equation}
\begin{equation}
     \lambda = \boldsymbol{\mathrm{Sigmoid}} \left(\mathbf{W_t} \cdot \mathbf{H_{\text{t}}} + \mathbf{W_c} \cdot \mathbf{H_{\text{c}}}\right),
\end{equation}
\begin{equation}
     \mathbf{H_{\text{f}}} = \left(1 - \lambda\right) \cdot \mathbf{H_{\text{t}}} + \lambda \cdot \mathbf{H_{\text{c}}},
\end{equation}
where $\mathbf{W^Q}$, $\mathbf{W^K}$, $\mathbf{W^V}$, $\mathbf{W_t}$ and $\mathbf{W_c}$ are learnable parameters.

Finally, we use the hidden states of the last layer from the fusion block as the input for the LLM to generate our prediction $G$. The scene graph generation loss is defined as follows:
\begin{equation}
\mathcal{L}_{SG}=-\sum_{l}^L \log P\left(g_l \mid g_{<l}, X_{\text{t}}; X_{\text{v}}\right),
\end{equation}
where $g$ is the generated token, $X_{\text{t}}$ is the text prompt containing $X_{\text{o}}$, and $X_{\text{v}}$ is the vision input.

\subsection{Scene Graph Triplets Selection}
\label{SGS}
After the scene graph generation stage, each image obtains a corresponding scene graph, consisting of a series of triplets. However, not all triplets are accurate, and some even present situations that contradict commonsense. Erroneous triplets not only provide no assistance to the generation process but may also introduce additional noise, causing adverse effects. To ensure more accurate and reasonable inference, the assisting triplets must be as precise as possible. Therefore, it is essential to select the high-quality generated triplets. Below are the scene graph selection methods we propose:

\textbf{Threshold Based Selection.} 
With the original image $I$ and the generated scene graph $G$, we use CLIP model to compute the normalized similarity score for each triplet and image region. The output scores are sorted in descending order, with higher scores representing higher similarity, which means they better fit the image scene. We set different thresholds for the selection process and then continuously select triplet with the highest confidence score from the ordered set. When the sum of accumulated confidence scores is less than the threshold, another triplet will be chosen until the threshold is reached. In order to find a better threshold, we conduct corresponding selection experiments, and the experimental results are shown in Section~\ref{ablation}. 

\begin{table*}[t!]
    \begin{center}
    \centering
        \scalebox{1}{
            \begin{tabular}{l|ccc|cccccccc|c}
                \toprule
                 & \multicolumn{3}{|c|}{\textbf{e-ViL Scores}} & \multicolumn{8}{c|}{\textbf{\textit{n}-gram Scores}} &  \multicolumn{1}{c}{\textbf{Learned Sc}}\\
                \cmidrule(lr){2-4} \cmidrule(lr){5-12} \cmidrule(lr){13-13} 
                
                \textbf{\textit{VCR}} & $S_O$ & $S_T$ & $S_E$ & B1 & B2 & B3 & B4 &  M & R-L & C &  S & BS \\
                \midrule
                e-UG                          & 19.3 & 69.8 & 27.6 & 20.7 &	11.6 & 	6.9 &	4.3 &	11.8 &	22.5 &	32.7 &	12.6 &	79.0   \\
                $\text{OFA-X}_{\text{VCR}}$     & 23.0 & \textbf{71.2} & 32.4 &	24.5 &	14.4 & 9.1 &6.1 &	12.2 &	25.1 &	48.5 &	18.8 &	79.8   \\
                $\text{OFA-X}_{\text{MT}}$     & 19.2 & 62.0 & 30.9 &	22.3 & 13.0 & 8.0 &	5.2 &	11.3 &	24.3 &	44.6 &	17.8 &	79.3   \\
                NLX-GPT                         & - & - & 32.6 & 24.7 &	15.0 &	9.6 &	6.6 &	12.2 &	26.4 &	46.9 &	18.8 &	80.3   \\
                $\text{UMAE}_{\text{VCR}}$      & 22.5 & 56.6 & 39.8 &	- &	- &	- &	12.3 &	16.7 &	28.9 &	48.2 & 	27.4 &	81.8 \\
                $\text{UMAE}_{\text{MT}}$      & 22.8 & 56.6 & 40.2 &	31.4 & 22.9 & 17.6 &	 \textbf{13.4} & 17.5 & 29.5 &	47.3 & 	26.5 & 	81.9 \\
                \textbf{G2(ours)}                            & \textbf{30.8} & 65.2 & \textbf{47.3} &	 \textbf{42.3} & \textbf{27.1} & \textbf{18.5} &13.3 &	 \textbf{21.9 }& \textbf{37.5 }&	 \textbf{57.7} & \textbf{28.3} & \textbf{91.1}  \\ 
                \midrule
                
                \multicolumn{13}{l}{\textbf{\textit{VQA-X}}} \\
                
                \midrule
                e-UG         & 36.4 & 80.5 & 45.3 &  57.3 & 42.7 & 31.4 & 23.2 & 22.1 &45.7 &  74.1 & 20.1 & 87.0   \\
                $\text{OFA-X}_{\text{MT}}$  & \textbf{45.5} & \textbf{92.6} & 49.2  & 64.0 & 49.4 & 37.6 & 28.6 & 23.1 & 51.0  & 110.2 & 22.6 & 86.8   \\
                NLX-GPT          & 40.6 & 83.0 & 49.0 & 64.2 & \textbf{49.5} & 37.6 & 28.5 & 23.1 & \textbf{51.5} & \textbf{110.6} & 22.1 & \textbf{86.9}   \\
                $\text{UMAE}_{\text{MT}}$   & 31.5 & 77.6  & 40.6 & 47.5 & 31.4 & 21.4 & 14.6 & 20.2 & 35.1 & 50.3 & 19.1 & 85.4 \\
                \textbf{G2(ours)}   & 45.2  & 90.1  & \textbf{50.3}  &	\textbf{66.3} & \textbf{49.5} & \textbf{38.1} & \textbf{28.8} & \textbf{24.0} & 49.4 &	 108.8 & \textbf{24.1} & 86.6 \\ 
                \midrule
                
                \multicolumn{13}{l}{\textbf{\textit{e-SNLI-VE}}} \\
                
                \midrule
                e-UG & 36.0 & \textbf{79.5} & 45.3 & 30.1	&19.9&	13.7&	9.6&	19.6&	27.8&	85.9&	34.5&	81.7   \\
                $\text{OFA-X}_{\text{MT}}$  & 35.6 & 78.9 & 45.1 &32.4 &	21.8 &	15.2 &	10.8 &	17.9 &	31.4 &	108.2 &	32.8 &	80.4   \\
                NLX-GPT                     & 34.6  & 73.9  & 46.9  & 37.0 &	25.3 &	17.9 &	12.7 &	18.8 &	34.2 &	\textbf{117.4} &	33.6 &	80.8   \\
                \textbf{G2(ours)}       & \textbf{45.3} & 75.2 & \textbf{60.2}  &	 \textbf{43.3} & \textbf{29.0} & \textbf{20.5} & \textbf{13.9} &	 \textbf{29.5 }& \textbf{41.6 } &	114.5 & \textbf{45.6} & \textbf{90.4}  \\
                \bottomrule
            \end{tabular}
        }
    \end{center}
    \caption{Filtered scores on VCR, VQA-X, and e-SNLI-VE dataset. B, M, R-L, C, S, and BS are abbreviations for BLEU, METEOR, ROUGE-L, CIDEr, SPICE, and BERTScore, respectively. \textbf{VCR} stands for training only on VCR, while \textbf{MT} represents the multitask setting. Note that the code for \cite{ref13} is not publicly available.}
    \vspace{-15pt}
    \label{experiment_results}
\end{table*}

\textbf{Confidence Score Based Selection.} 
Although threshold-based selection can help select higher-quality scene graphs to some extent, it remains challenging to choose a better scale, even with constant refinement of thresholds. Therefore, we regard that it as a better approach, allowing the model to select more effective scene graphs by itself during training. Thus, we propose the confidence score based selection method, allowing the confidence scores to guide the model in scene graph selection. We first input each triplet from the generated scene graph along with the image $I$ into the CLIP model to obtain their respective confidence scores: 
\begin{equation}
    c_k = \boldsymbol{\mathrm{CLIP}}(I, \tau_k), \quad \forall \tau_k \in G,
\end{equation}
where $\tau_k$ is the \emph{k}-th generated triplet in scene graph $G$ and $c_k$ is the corresponding confidence score. Upon acquiring the normalized confidence scores for the scene graph from the CLIP, these values are subsequently expanded to match the dimensions of the text input embedding. More precisely, the segment of input tokens associated with each scene graph triplet consists of identical corresponding scores, whereas the segments external to the scene graph are assigned a default value of 1. Then, the confidence-weighted text representation is obtained by the following function:
\begin{equation}
\alpha_{ij} = 
\begin{cases} 
\frac{e_{ij} c_{ij}}{\sum\limits_{j} e_{ij} c_{ij}}, & \text{if } c_{ij} \text{ in } G, \\
e_{ij}, & \text{otherwise},
\end{cases}
\end{equation}
where $e_{ij}$ represents the \emph{i}-th token of \emph{j}-th input  , $c_{ij}$ denotes the confidence score generated by CLIP with the image and the corresponding triplet, and $\alpha_{ij}$ is corresponding to the \emph{j}-th weighted input. The $\alpha_{ij}$, encapsulating confidence scores, facilitates the provision of insights regarding scene graph quality throughout the training process. This enables the model to focus more on triplets with higher confidence while concurrently reducing the likelihood of selecting those with lower scores.  As a result, the model independently acquires the capacity to opt for higher-quality scene graphs as its foundation for explanation.

\subsection{VCR Answer and Explanation Generation}
\label{vcrg} 
In the VL-NLE setup, the focus of VCR shifts from multiple-choice problem-solving to a generative approach, which aims to generate corresponding answers and explanations. We combine these two discrete tasks into a single one, generating both the answer and explanation simultaneously. As a result, we use the text prompt: ``[\emph{Context:$\{G\}$ Question:$\{Q\}$}]'' as our text input $X_{\text{t}}$. In detail, $G=\{sub_1\ obj_1\ rel_1;\ \dots\ ; sub_{\sigma}\ obj_{\sigma}\ rel_{\sigma}\}$, which is the scene graph triplets generated in Section~\ref{sgg}. The question $Q$ is represented as $Q=\{q_1, q_2, \dots\ , q_h\}$, $h$ is the length of the input question. For visual input $X_{\text{v}}$, we use the same patch sequence as in the scene graph construction. We also apply a single-head cross-modal attention and gated mechanism for modality fusion. During the training and inference process, as described in Section~\ref{SGS}, the confidence scores of the scene graph triplets are combined with the input embedding and participate in the following attention computation. In this way, the model is able to learn how to select more accurate scene graph triplets without increasing computational complexity. Meanwhile, we combine the ground truth answer and explanation in the form of ``[\emph{$\{A\}$ Because:$\{R\}$}]'' as our training label for $(Q \rightarrow AR)$ generation. Considering the original design of VCR, which encompasses two sub-tasks, $(Q \rightarrow A)$ and $(QA \rightarrow R)$, we also provide corresponding prompt templates. Be more specific, we provide ``[\emph{Context:$\{G\}$ Question:$\{Q\}$ Answer:}]'' for $(Q \rightarrow A)$ sub-task and ``[\emph{Context:$\{G\}$ Question:$\{Q\}$ Answer:$\{A\}$ Because:}]'' for $(QA \rightarrow R)$ sub-task. Their corresponding labels are ``[\emph{Answer:$\{A\}$}]'' and ``[\emph{Explanation:$\{R\}$}]'' respectively. The loss is defined as the following cross-entropy loss:
\begin{equation}
\mathcal{L}_E=-\sum_{i}^I \log P\left(y_i \mid y_{<i}, X_{\text{t}}; X_{\text{v}}\right),
\end{equation}
where $y$ is the generated token, $X_{\text{t}}$ is the text input containing $Q$ and $G$, and $X_{\text{v}}$ is the vision input.


\section{Experimental Setup}
\label{others}

\subsection{Datasets}
\textbf{Visual Genome (VG).} 
VG \cite{ref5} dataset is the most frequently used scene graph dataset. It contains over 100K images, where each image has an average of 21 objects, 18 attributes, and 18 relationships. Its intention is to Promote research in advanced semantic understanding of images.

\textbf{Visual Commonsense Reasoning (VCR).} 
VCR \cite{ref1} is a visual understanding dataset, including 290k pairs of questions, answers, and rationales, over 110k unique movie scenes. The intention is to uncover the deep relationships among the objects and to correctly choose answer and explanation. Under the NLE setting, we adopt the changed VCR dataset provided by NLX-GPT \footnote{https://github.com/fawazsammani/nlxgpt}. The training, validation, and testing split consist of 191,657, 21,326, and 26,534 samples respectively. Each data sample consists of a question, an image and a  pair of ground truth answer and explanation.

\textbf{VQA-X.} 
The VQA-X \cite{park2016attentive} dataset contains human written explanations for a subset of questions from the VQA v2 dataset \cite{goyal2017making}. The image-question pairs are split into train, dev, and test with 29.5k, 1.5k, and 2k instances, respectively. The original task is formulated as a multi-label classification task of 3,129 different classes. Under the NLE setting, this task is turned into a generation task, with each data sample consists of a question, an image and several pairs of ground truth answer and explanation.

\begin{table}[t!]
    \begin{center}
    \centering
        \scalebox{1.1}{
            \begin{tabular}{l|c|ccc}
                \toprule
                Method  &BBox & R@20 & R@50 & R@100 \\
                \midrule
                SGTR   & \Checkmark &23.62 &30.38 &34.85 \\
                Pix2SG & \XSolidBrush &21.51 &24.81 &26.66 \\
                \textbf{ours}   & \XSolidBrush &23.22 &29.93 &44.76 \\
                \bottomrule
            \end{tabular}
        }
  \end{center}
  \caption{\textbf{LF-SGG results of current models at R@k on Visual Genome dataset.}}
  \vspace{-20pt}
  \label{scenegraph}
\end{table}

\textbf{e-SNLI-VE.} 
The e-SNLI-VE \cite{ref12} dataset contains over 430k instancess, the currently largest dataset
for VL-NLE. It is constructed by merging the explanations from e-SNLI \cite{camburu2018snli} and  the image-sentence pairs from SNLI-VE \cite{xie2019visual}. The validation and test sets were relabelled by hand.

\subsection{Baseline Methods}
We compare our method to the following baselines: 

\textbf{e-UG.} 
e-UG \cite{ref12} Combines GPT-2 with UNITER, using visual features of regions with Faster R-CNN, and encodes location features as vision input. With the embeddings of the image regions and question words prepended to the textual question and predicted answer, it leverages UNITER's contextualized embeddings to condition GPT-2.

\textbf{OFA-X.} 
OFA-X \cite{ref15} takes the OFA \cite{wang2022ofa} model as the backbone, which is a standard encoder-decoder Transformer architecture. Prompting the model with the question followed by four different choices, it chooses the right answer and explanation by their possibilities.

\textbf{NLX-GPT.} 
NLX-GPT \cite{ref14} is an encoder-decoder model, which is composed of a visual backbone that encodes the image and a distilled GPT-2. It first pretrains the Distilled GPT-2 on image captioning, then fine-tunes it on VL-NLE datasets.

\textbf{UMAE.} 
UMAE \cite{ref13} is also based on the OFA model. It designs corresponding prompts for multiple datasets and conducts multi-task training on VL-NLE tasks.


\subsection{Evaluation Metrics}
To assess the quality of the generated scene graphs and ascertain their contribution to the generation of explanations, we conduct evaluations separately on scene graph generation and VCR answer and explanation generation under their corresponding metrics.

For location-free scene graph generation, we adopt Heuristic Tree Search (HTS) \cite{ref2} to evaluate generation quality, which is used to calculate the degree of overlap between graphs and obtain recall scores. The generated scene graphs are presented in the form of sequences during the prediction stage. We first convert the predicted sequences and label sequences into graph structure, and then use this tree-based search algorithm to calculate Recall@20, 50, and 100.

For VCR answer and explanation generation, we adopt different evaluation methods, including automatic evaluation and human evaluation. For automatic evaluation, we use metrics for natural language generation: BLEU \cite{ref26}, METEOR \cite{ref28}, ROUGE-L \cite{ref27}, CIDEr \cite{ref29}, SPICE \cite{ref30}, and BERTScore F1 \cite{ref31}. We also use the e-ViL score \cite{ref12} for evaluation. The e-ViL score consists of $S_T$ , $S_E$, and $S_O$, which represent task accuracy, n-gram explanation score, and the total score (product of $S_T$ and $S_E$), respectively. $S_E$ is obtained by calculating the harmonic mean of ROUGE-L, METEOR, CIDEr, SPICE, and additional BERTScore. To compare our method with previous works \cite{ref9, ref13, ref14, ref15}, we similarly assess the performance of our model using filtered generation results. During the filtering process, we employ a BERTscore threshold of 0.92 to determine the correctness of the answers. Answers with a BERTscore below 0.92 are considered incorrect, while those reaching 0.92 are deemed correct and further evaluated for their explanations. Furthermore, we also adopt human evaluation, and the method is identical to \cite{ref12}: Randomly selecting 300 examples with correct answers from the test set, and the annotators need to choose one from (yes, weak yes, weak no, and no) as a response for whether the explanation justifies the answer. It should be noted that we only evaluate the explanations with correct answers.

\begin{table}[t!]
    \begin{center}
    \centering
        \scalebox{1}{
            \begin{tabular}{l|ccccccc}
                \toprule
                Method  & B4 &  M & R-L & C &  S & BS \\
                \midrule
                G2 ($w/o$ SG)    &	11.8 &	19.7 &	36.0 &	51.7 &	24.8 & 89.9   \\
                G2 (0.7)         &	12.3 &	20.5 &	36.6 &	53.3 &	26.5 &	90.3   \\
                G2 (0.8)        &	13.1&	21.7&	37.0&	56.1&	27.9&	90.5\\
                G2 (0.9)        &	12.4 &	20.6 &	36.8 &	52.1 &	26.5 &  89.9  \\
                G2 (automatic)  & \textbf{13.3}&\textbf{21.9}& \textbf{37.5}& \textbf{57.7}& \textbf{28.3}& \textbf{91.1}  \\ 
                \bottomrule
  \end{tabular}
    }
    \end{center}
     \caption{\textbf{Ablation study on using varying scene graph selection methods.}  \textbf{0.7/0.8/0.9} represents different manually selection thresholds. \textbf{Automatic} means our automatic selection method.}
    \vspace{-20pt}
    \label{filter}
\end{table}

\subsection{Implementation Details}

We use 8 × NVIDIA 3090 24Gb for our experiments. For scene graph generation, we train models for 25K steps, with a batch size of 8. For explanation generation, we train models for 20K steps, with a batch size of 16. We both use AdamW as our optimizer, and the learning rate is $5e^{-5}$. The warm-up steps are 1K and 2K, respectively, and the trained models are both evaluated every 5,000 steps. We chose the best model for evaluation. We use the patch sequence generated by CLIP as our vision input, and we initialize our model using the Llama-3.2-1B \footnote{https://huggingface.co/meta-llama/Llama-3.2-1B}.


\section{Results and Analysis}

\subsection{Results of Scene Graph Constructing}
Unlike most SSG models, our method adopts a location-free setting, where explicit location information is not provided. Consequently, we only compare our model with those under the same configuration, and the results are shown in Table~\ref{scenegraph}. Pix2SG \cite{ref2} is the earliest and currently the only model proposing SGG in a location-free setting. It is Based on the Pix2Seq backbone \cite{ref60}, which is a multi-modal model capable of representing the object detection task as sequence prediction. Our framework does not incorporate location information, resulting in a small amount of noise during generation. Consequently, its performance appears to be rather mediocre in Recall@20. However, the text prompts we provided are able to offer information hints and G2 surpasses Pix2SG in R@50 and R@100. 

\subsection{Results of Answer and Explanation Generation}


\begin{figure}[t!]
    \centering
    \includegraphics[width=0.85\columnwidth]{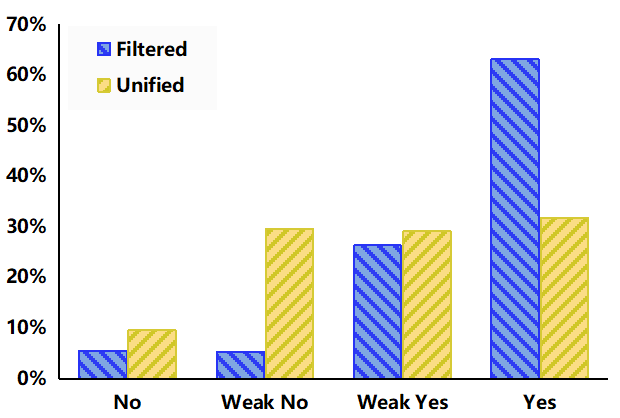}
    \caption{\textbf{Human evaluation of the ground-truth explanations for the VCR dataset.} \textbf{Filtered} refers to the evaluation with correct answers. \textbf{Unified} refers to the assessment of both answers and explanations.}
    \Description{..}
    \label{Human_evaluation}
\end{figure}

\textbf{VCR.}  
Table~\ref{experiment_results} shows that previous work (e.g., e-UG, OFA-X, and NLX-GPT) achieves similar performances on VCR. They display a lower score on n-gram metrics like BLEU and SPICE. UMAE obtains significant performances compared with previous ones, including an increase in BLEU-4 from $6.6\%$ to $13.4\%$. However, compared with other models, it experiences a decrease of about 10\% in score on the $S_T$ metric. This suggests that there are deficiencies in the accuracy of his generated results.

Our proposed method achieves impressive performance comparably to previous work. Specifically, after employing the confidence score based selection method, G2 is able to reach scores of 21.9 on METEOR, 37.5 on ROUGE-L, 57.7 on CIDEr, and 91.1 on BERTScore, respectively. These findings further illustrate that patch sequence and object relationships provided by scene graphs aid in generating more reasonable explanations. At the same time, on the e-ViL metrics, we achieve better scores on $S_O$ and $S_E$, but the performance on $S_T$ does not surpass the existing SOTA. In order to better evaluate the generation quality, this paper randomly select test samples from the test set that are filtered and unfiltered by BERTScore for human evaluation, which is shown in Figure~\ref{Human_evaluation}. 63.1\% of the filtered examples are considered to have well demonstrated and explained the answer part, and only 5.4\% of them fail entirely in reasoning and explanation. This indicates that from the perspective of human cognition, the vast majority of samples are in line with commonsense and are explanatory. This finding suggests that, despite the discrepancies observed between generated sentences and ground truth during the automatic evaluation process, the annotators regard these sentences as accurate and, in some instances, superior to the ground truth. In the unfiltered set, the proportions of ``weak no'' and ``weak yes'' increases, while the proportion of ``no'' is still small. This also shows that the quality of reasoning can maintain stability even when the answer is not completely correct.

\textbf{VQA-X.}  
As shown in Table~\ref{experiment_results}, our framework achieves relatively high scores in n-gram and BERTScore metrics. Among these, METEOR and SPICE receive the highest scores of 24.0 and 24.3 and surpass previous work in $S_E$ scores. However, the answers and explanation sentences provided in the VQA-X dataset are relatively short (usually composed of words or phrases), while our proposed framework tends to generate more extended and detailed explanations, resulting in a reduced overlap with the ground truth, which will indirectly lead to a decrease in these evaluation scores. Therefore, the method proposed in this paper is slightly lower than the baseline model in individual indicators. However, from a comprehensive perspective, G2 still demonstrates good performance.


\begin{table}[t!]
    \begin{center}
    \centering
        \scalebox{1}{
            \begin{tabular}{l|c|cccc}
                \toprule
                Method & Dimension & R@20 & R@50 & R@100 \\
                \midrule
                DETR                              & (100, 256) &20.62 & 27.01 & 43.86  \\
                $\text{CLIP}_{\text{global}}$     & (1, 512) &  10.26 & 15.82 & 27.56  \\
                $\text{CLIP}_{\text{patch}}$      & (49, 2048) & \textbf{23.22} &\textbf{29.93} &\textbf{44.76}  \\
                \bottomrule
            \end{tabular}
        }
  \end{center}
  \caption{Ablation study on using different vision features for scene graph generation on VG dataset.}
  \vspace{-20pt}
  \label{vision_feature}
\end{table}


\begin{figure*}[!t]
  \centering
  \includegraphics[width=0.95\textwidth]{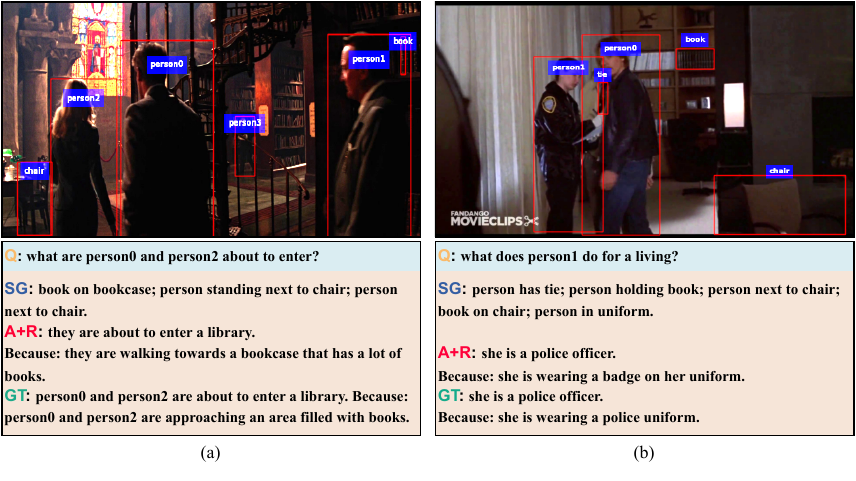}
  \caption{\textbf{Case study of G2 on VCR dataset.} ``Q'', ``SG'', ``A+R'', and ``GT'' denote the question, generated scene graph, predictive answer and rationale, and ground truth answer \& explanation, respectively.}
  \Description{..}
  \label{case1}
\end{figure*}

\textbf{e-SNLI-VE.}  
The experimental results on the e-SNLI-VE dataset also exhibit the effectiveness of our proposed framework. According to the results in Table~\ref{experiment_results}, G2 obtains relatively high scores across all metrics, particularly surpassing previous work in metrics such as METEOR and ROUGE-L. Additionally, compared to prior work, it achieves a 7\% improvement in the comprehensive score $S_O$. This further corroborates that our proposed framework is capable of generating explanations that conform to the scene information, ultimately allowing for a more precise determination of whether the textual description aligns with the image.

\subsection{Ablation Analysis}
\label{ablation}

\textbf{Different Vision Features.} 
We perform the ablation study to figure out the effect of using different vision features. The results are displayed in Table~\ref{vision_feature}. DETR \cite{ref66} is an end-to-end Transformer-based model for object detection. CLIP is used to extract patch features and the global feature separately. It can be observed that DETR only shows a slight lag on these metrics compared to the patch sequence. Meanwhile, the encoded global feature shows obvious shortcomings in Recall. We conjecture that this is because of the compression during encoding, which causes the information loss, while the patch feature can provide more complete visual information for LLMs.

\textbf{Different Scene Graph Selection Methods.} 
We also conduct ablation experiments on different scene graph selection methods. The results are displayed in Table~\ref{filter}. It can be observed that when the threshold is set to 0.8, the filtered results can achieve optimal performance. However, when the threshold is reduced or increased, the performance shows a declining trend, even falling below the level of not using scene graphs. According to this result, it can be seen that the quality of the generated scene graphs is not consistent. Therefore, providing fewer scene graphs will result in a loss of information, while providing more scene graphs of poor quality will result in an increase in noise. Compared with the threshold-based selection method, the overall performance of confidence score based selection is much better. This indicates that the confidence score based selection method can avoid selecting a suitable threshold and better assist the model in selecting effective scene graphs.


\subsection{Case Study}
Finally, we present the case study and provide analysis for cases, so that we can have a clear insight into the proposed approach. Our visualization results on the VCR dataset are shown in Figure~\ref{case1}. It can be seen that the sequence of scene graph generated by G2 provides detailed relationship information among essential objects.  In the case of Figure~\ref{case1}~(a), the generated ``\emph{book on bookcase}'' indicates that this is in a library, and G2 correctly applies it to generate the explanation.
The example shown in Figure~\ref{case1}~(b) also demonstrates a stronger reasoning ability of our model. With the generated ``\emph{person has tie}'' and ``\emph{person in uniform}'', G2 generates an explanation that closely matches the ground truth. It is also worth noting that the generated rationale mentions the ``\emph{badge}'', an object not referred to in the scene graph, but is very important for inference. This shows that the scene graph and the model's scene understanding can complement each other, jointly enhancing the credibility of the explanation. More cases can be seen in the Appendix.

\section{Conclusion}
In this work, we formally study the problem of VCR explanation generation. To address the issue of lacking perception of scene relationships in a commonsense reasoning process, we propose a scene graph-based generation framework. The decoupled framework separates the scene graph generation and explanation generation into two stages. We show that generated scene graphs help produce more reasonable rationales that contribute to more accurate answers. Consequently, our approach achieves impressive performance comparably to other advanced methods. We also conduct ablation studies to verify the effectiveness of the automatic scene graph selection method on the VCR benchmark. We will explore more comprehensive visual information mining methods and enhance the application of logical reasoning methodologies in future work.

\section{Acknowledgement}
This research is supported by the National Natural Science Foundation of China (No.62476127, No.62106105),  the Natural Science Foundation of Jiangsu Province (No.BK20242039), the CCF-Baidu Open Fund (No.CCF-Baidu202307), the CCF-Zhipu AI Large Model Fund (No.CCF-Zhipu202315), the Fundamental Research Funds for the Central Universities (No.NJ2023032), the Scientific Research Starting Foundation of Nanjing University of Aeronautics and Astronautics (No.YQR21022), and the High Performance Computing Platform of Nanjing University of Aeronautics and Astronautics.

\balance

\bibliographystyle{ACM-Reference-Format}
\bibliography{sample-base}

\appendix

\section{Visualization cases}
\begin{figure*}[htbp]
  \centering
  \includegraphics[width=0.8\textwidth]{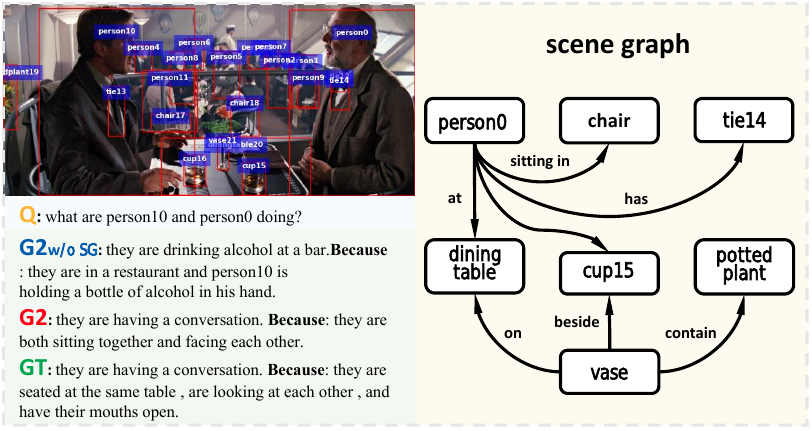}
  \caption{\textbf{Representative visualization cases of the proposed \textit{\textbf{G2}}.} ``G2 w/o SG'', ``G2'', and ``GT'' denote the answer and explanation of \textit{\textbf{G2}} without scene graph, \textit{\textbf{G2}}, and ground truth respectively.}
  \Description{..}
  \label{appendix_1}
\end{figure*}

\begin{figure*}[htbp]
  \centering
  \includegraphics[width=0.8\textwidth]{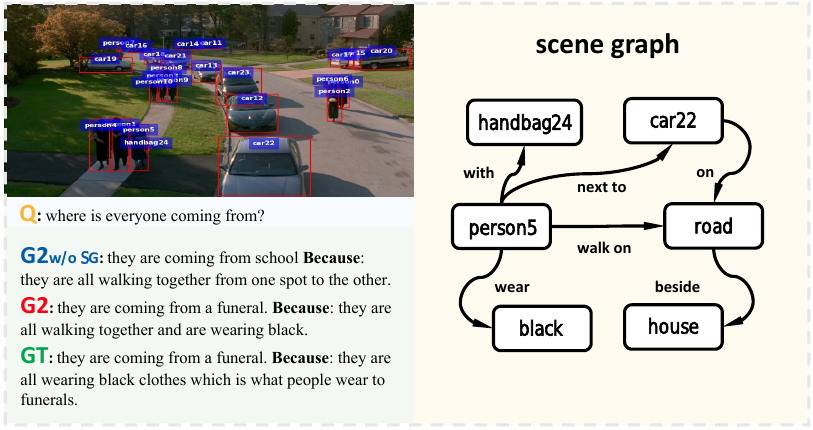}
  \caption{\textbf{Representative visualization cases of the proposed \textit{\textbf{G2}}.} ``G2 w/o SG'', ``G2'', and ``GT'' denote the answer and explanation of \textit{\textbf{G2}} without scene graph, \textit{\textbf{G2}}, and ground truth respectively.}
  \Description{..}
  \label{appendix_2}
\end{figure*}

\begin{figure*}[htbp]
  \centering
  \includegraphics[width=0.8\textwidth]{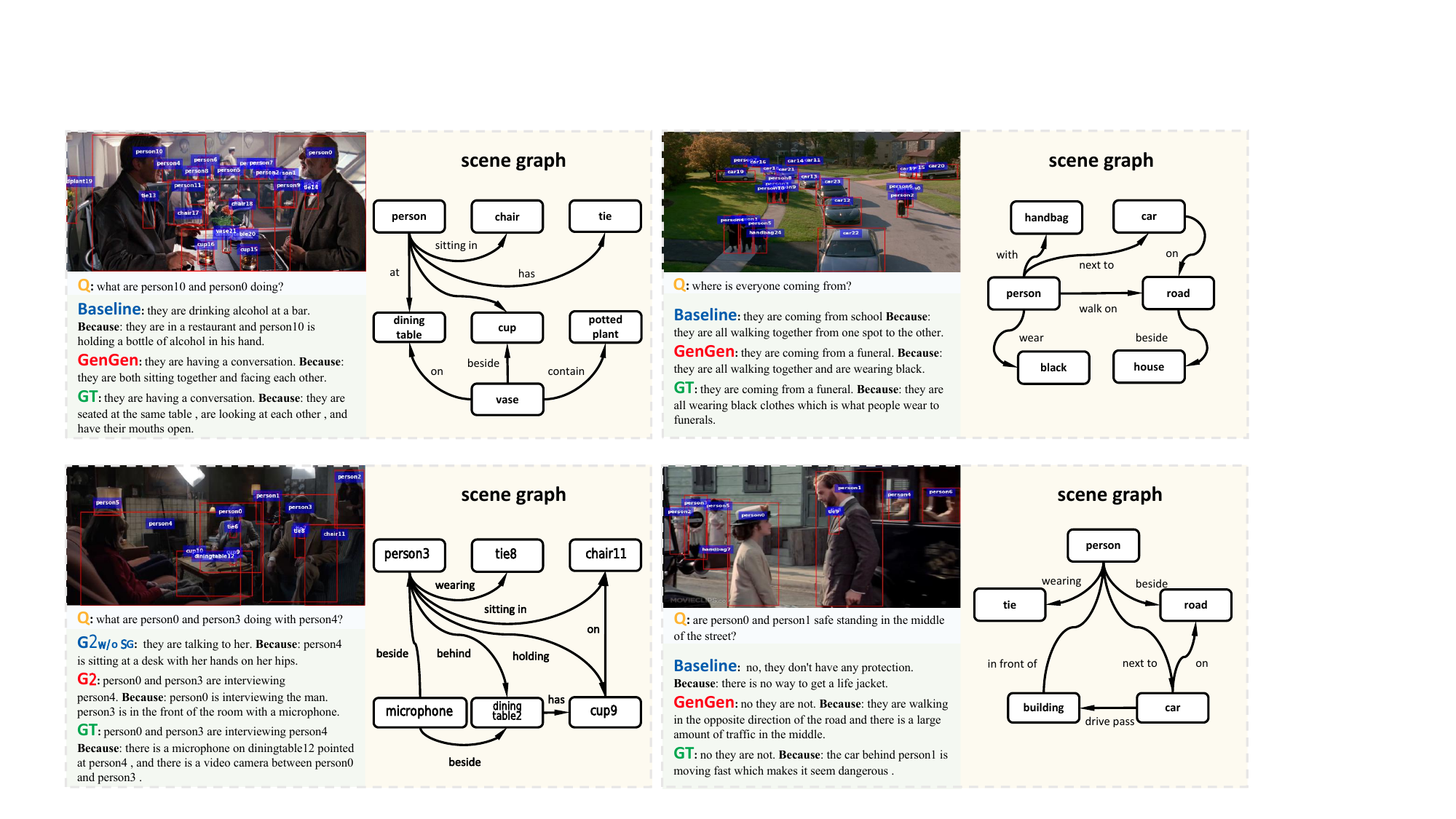}
  \caption{\textbf{Representative visualization cases of the proposed \textit{\textbf{G2}}.} ``G2 w/o SG'', ``G2'', and ``GT'' denote the answer and explanation of \textit{\textbf{G2}} without scene graph, \textit{\textbf{G2}}, and ground truth respectively.}
  \Description{..}
  \label{appendix_3}
\end{figure*}

\begin{figure*}[htbp]
  \centering
  \includegraphics[width=0.8\textwidth]{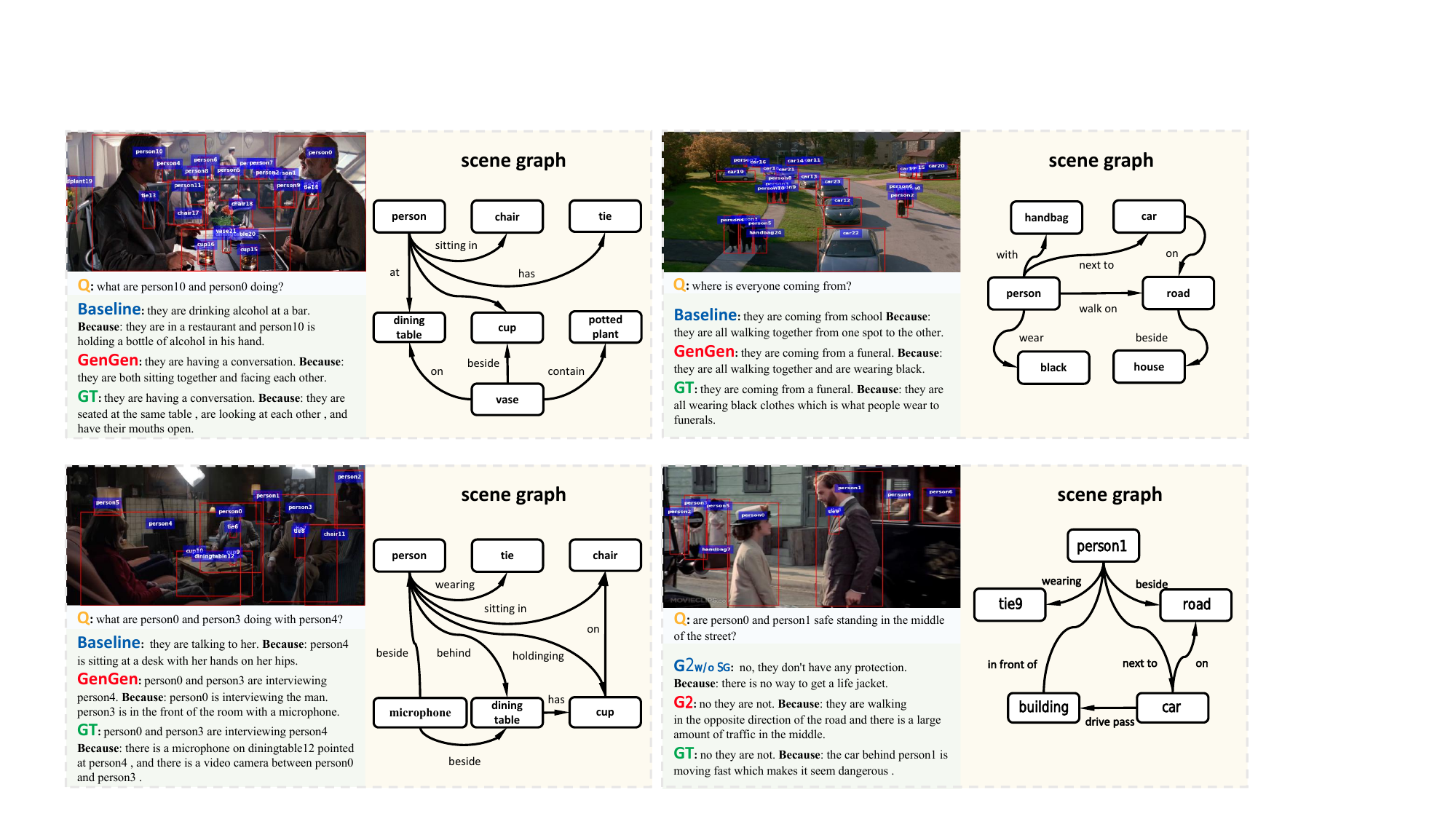}
  \caption{\textbf{Representative visualization cases of the proposed \textit{\textbf{G2}}.} ``G2 w/o SG'', ``G2'', and ``GT'' denote the answer and explanation of \textit{\textbf{G2}} without scene graph, \textit{\textbf{G2}}, and ground truth respectively.}
  \Description{..}
  \label{appendix_4}
\end{figure*}

We showcase more visualization cases in Figure~\ref{appendix_1}, ~\ref{appendix_2}, ~\ref{appendix_3}, and~\ref{appendix_4}.

\end{document}